\documentclass[a4paper,twoside]{article}

\usepackage{epsfig}
\usepackage{subcaption}
\usepackage{calc}
\usepackage{algorithm2e}
\usepackage{amssymb}
\usepackage{amstext}
\usepackage{amsmath}
\usepackage{amsthm}
\usepackage{multicol}
\usepackage{pslatex}
\usepackage{apalike}
\usepackage{algorithmic}
\usepackage{graphicx}
\usepackage{textcomp}
\usepackage{xcolor}
\usepackage{tikz}
\usepackage[export]{adjustbox}
\usetikzlibrary{quantikz2}
\usepackage{braket}
\usepackage{cleveref}
\usepackage[bottom]{footmisc}
\usepackage{SCITEPRESS}     

\begin{document}

\title{Illustration of Barren Plateaus in Quantum Computing}

\author{\authorname{Gerhard Stenzel\sup{1}\orcidAuthor{0009-0009-0280-4911}, Tobias Rohe\sup{1}, Michael Kölle\sup{1}, Leo Sünkel\sup{1}, Jonas Stein\sup{1}, Claudia Linnhoff-Popien\sup{1}}
\affiliation{\sup{1}LMU Munich}
\email{gerhard.stenzel@ifi.lmu.de}
}


\abstract{    Variational Quantum Circuits (VQCs) have emerged as a promising paradigm for quantum machine learning in the NISQ era. While parameter sharing in VQCs can reduce the parameter space dimensionality and potentially mitigate the barren plateau phenomenon, it introduces a complex trade-off that has been largely overlooked. This paper investigates how parameter sharing, despite creating better global optima with fewer parameters, fundamentally alters the optimization landscape through deceptive gradients—regions where gradient information exists but systematically misleads optimizers away from global optima. Through systematic experimental analysis, we demonstrate that increasing degrees of parameter sharing generate more complex solution landscapes with heightened gradient magnitudes and measurably higher deceptiveness ratios. Our findings reveal that traditional gradient-based optimizers (Adam, SGD) show progressively degraded convergence as parameter sharing increases, with performance heavily dependent on hyperparameter selection. We introduce a novel gradient deceptiveness detection algorithm and a quantitative framework for measuring optimization difficulty in quantum circuits, establishing that while parameter sharing can improve circuit expressivity by orders of magnitude, this comes at the cost of significantly increased landscape deceptiveness. These insights provide important considerations for quantum circuit design in practical applications, highlighting the fundamental mismatch between classical optimization strategies and quantum parameter landscapes shaped by parameter sharing.}

\onecolumn \maketitle \normalsize \setcounter{footnote}{0} \vfill
\section{INTRODUCTION}\label{sec:intro}

The emergence of quantum computing has catalyzed the development of Quantum Machine Learning (QML), an interdisciplinary field that leverages quantum phenomena to potentially overcome classical computational limitations~\cite{Biamonte2017QuantumLearning,Schuld2019EvaluatingHardware}. In the current Noisy Intermediate-Scale Quantum (NISQ) era, hybrid quantum-classical approaches utilizing Variational Quantum Circuits (VQCs) have become the predominant implementation paradigm~\cite{Preskill2018QuantumBeyond,Cerezo2021VariationalAlgorithms}.

These VQCs function as quantum analogues to classical neural networks, employing parameterized quantum gates that are optimized by classical routines~\cite{Benedetti2019ParameterizedModels,McClean2016TheAlgorithms}. However, VQC optimization faces significant challenges, particularly the barren plateau phenomenon, where gradients vanish exponentially with increasing system size~\cite{McClean2018BarrenLandscapes}.

Parameter sharing in VQCs has emerged as a promising strategy to address these challenges. By reducing the parameter space dimensionality and enforcing useful symmetries, it offers potential advantages in optimization efficiency and generalization capabilities~\cite{Skolik2021LayerwiseNetworks}. Our research confirms that parameter sharing can indeed lead to superior global optima while using fewer parameters - a seemingly ideal solution for quantum circuit design.

However, this approach introduces a critical trade-off that has been largely overlooked: parameter sharing creates complex dependencies in gradient information that can fundamentally alter the optimization landscape. Our work identifies a complementary challenge to barren plateaus: deceptive gradients - regions where gradient information exists but systematically misleads optimizers away from global optima. This deceptiveness emerges from the interplay of parameter interdependencies and quantum-specific effects with no classical analogue~\cite{Arrasmith2021EffectOptimization,Cerezo2021CostCircuits}.

This research investigates the fundamental question: while parameter sharing in VQCs can create better global optima with fewer parameters, what is the cost in terms of landscape deceptiveness and practical trainability?

In \cref{sec:related}, we discuss the related work regarding the intricate difficulties when optimizing variational quantum circuits, particularly focusing on the challenges posed by barren plateaus (\cref{sec:related:barren}) and deceptive gradients (\cref{sec:related:deceptive}).

In \cref{sec:method}, we introduce our methodology, which includes the concept of resolution (\cref{sec:resolution}), the definition of deceptiveness (\cref{sec:deceptiveness}), and our experimental design for comparing optimizer performance across different parameter sharing configurations (\cref{sec:experiments}).

In \cref{sec:results}, we present our experimental findings, which include a comprehensive analysis of optimizer trajectories under various parameter sharing configurations (\cref{sec:trajectories}). We subsequently provide a thorough evaluation of optimizer success rates across a spectrum of learning rate parameters (\cref{sec:successrate}), followed by a comparative assessment of gradient-based optimization methodologies contextualized within our quantitative framework of landscape deceptiveness. These analyses collectively elucidate the intricate relationship between parameter sharing strategies and optimization efficacy in the quantum circuit domain.

We summarize and discuss our results in \cref{sec:conclusion}, where we also outline potential future work. Our appendixes, starting with \cref{sec:defaultcircuit}, provide additional insights and detailed analyses related to our methodology and results.

Our contributions can be summarized as follows:
\begin{itemize}
    \item We demonstrate a fundamental trade-off in quantum circuit design: parameter sharing significantly enhances circuit expressivity and improves global minima by orders of magnitude, while simultaneously increasing landscape deceptiveness.
    
    \item We quantify how increasing degrees of parameter sharing generate more complex solution landscapes with heightened gradient magnitudes, resulting in measurably higher deceptiveness ratios.
    
    \item We develop and validate a novel gradient deceptiveness detection algorithm that efficiently identifies misleading regions in quantum optimization landscapes, with evidence that deceptiveness ratio remains largely independent of sampling resolution.
    
    \item We provide empirical evidence that traditional gradient-based optimizers (Adam, SGD) show progressively degraded convergence as parameter sharing increases, revealing a fundamental mismatch between classical optimization strategies and quantum landscapes.
    
    \item We establish that optimizer hyperparameter selection (particularly learning rate) becomes increasingly critical in highly deceptive landscapes, with direct implications for practical quantum circuit training.
    
    \item We introduce a quantitative methodological framework for measuring optimization difficulty in quantum circuits, valuable for evaluating circuit design choices and their impact on trainability.
\end{itemize}

\section{RELATED WORK}\label{sec:related}

\subsection{Classical Optimization Algorithms}\label{sec:related:classical}

Optimization algorithms form the backbone of machine learning, enabling the training of models by minimizing objective functions. Among these, gradient-based methods such as Stochastic Gradient Descent (SGD) \cite{Robbins1951AMethod}, momentum-based methods \cite{Polyak1964SomeMethods}, and adaptive learning rate methods like AdaGrad \cite{Duchi2011AdaptiveOptimization}, RMSProp \cite{Tijmen2012LectureMagnitude}, and Adam \cite{Kingma2014Adam:Optimization} have become particularly prominent due to their efficiency and scalability.

While these classical optimization algorithms have proven highly effective for traditional machine learning models, they face unique challenges when applied to quantum machine learning, particularly in the context of variational quantum circuits, as we will explore in subsequent sections.

\subsection{Quantum Machine Learning Fundamentals}\label{sec:related:quantum}

Quantum Machine Learning represents the intersection of quantum computing and machine learning, aiming to leverage quantum mechanical phenomena to enhance learning capabilities \cite{Biamonte2017QuantumLearning}. Unlike classical machine learning, which operates on classical bits, QML utilizes quantum bits or qubits that can exist in superpositions of states, potentially offering computational advantages for specific problems \cite{Schuld2015AnLearning}.

The field of QML can be approached from two complementary directions: using quantum computers to accelerate classical machine learning algorithms, and applying machine learning techniques to quantum systems \cite{Wittek2014QuantumMining}. In the NISQ (Noisy Intermediate-Scale Quantum) era \cite{Preskill2018QuantumBeyond}, characterized by limited qubit counts and high error rates, hybrid quantum-classical approaches have emerged as the most practical implementation strategy. These methods execute certain parts of algorithms on quantum processors while performing others on classical computers \cite{Cerezo2021VariationalAlgorithms}. General pipelines for solving optimization problems on quantum hardware have been developed to systematize this process \cite{Rohe2024FromHardware}.
Recent work also explores hybrid quantum approaches for sequence generation and for accelerating state vector simulation through gate-matrix caching and circuit splitting \cite{qml2,qml5}.

Central to many QML implementations are variational quantum circuits, which serve as quantum analogues to classical neural networks \cite{Benedetti2019ParameterizedModels}. These circuits consist of parameterized quantum gates whose parameters are optimized using classical optimization routines. The quantum circuit prepares a quantum state that encodes the solution to a problem, and measurements of this state provide the output used to compute a cost function \cite{Schuld2020Circuit-centricClassifiers}.

Data encoding represents another fundamental challenge in QML, as classical data must be mapped into quantum states through various embedding strategies. These include basis encoding, amplitude encoding, and angle encoding, each with different resource requirements and expressivity characteristics \cite{Havlicek2019SupervisedSpaces}. The choice of encoding strategy significantly impacts the potential quantum advantage and the trainability of the resulting model.

Despite theoretical promise, QML faces several practical challenges, including hardware limitations, noise susceptibility, and the difficulty of loading classical data efficiently into quantum states \cite{Aaronson2015ReadPrint}. Additionally, the optimization of variational quantum circuits presents unique obstacles not encountered in classical machine learning, such as barren plateaus and deceptive gradients, which we will examine in detail in later sections.

\subsection{Variational Quantum Circuits with Parameter Sharing}\label{sec:related:variational}

Variational Quantum Circuits (VQCs), also known as Parameterized Quantum Circuits (PQCs), have emerged as the cornerstone of quantum machine learning in the NISQ era \cite{Cerezo2021VariationalAlgorithms}. These circuits consist of parameterized quantum gates whose parameters are tuned through classical optimization to minimize a cost function. A typical VQC workflow involves preparing an initial state, applying a parameterized unitary transformation, and measuring an observable to compute the cost function \cite{McClean2016TheAlgorithms}.

The design of VQCs presents a fundamental trade-off between expressivity and trainability. Highly expressive circuits can represent complex functions but often suffer from optimization difficulties \cite{Holmes2022ConnectingPlateaus}. Parameter sharing has emerged as a promising technique to navigate this trade-off by reducing the number of independent parameters while maintaining sufficient expressivity \cite{Sim2019ExpressibilityAlgorithms}.

Parameter sharing in VQCs involves constraining certain parameters to have identical values, analogous to weight sharing in classical convolutional neural networks \cite{Cong2019QuantumNetworks}. This approach offers several advantages: it reduces the dimensionality of the optimization landscape, decreases the number of parameters that need to be optimized, and can improve generalization by enforcing symmetries in the circuit \cite{Skolik2021LayerwiseNetworks}.

Self et al. \cite{Self2021VariationalSharing} demonstrated that parameter sharing can be leveraged to solve related variational problems in parallel through their Bayesian Optimization with Information Sharing (BOIS) approach. By sharing quantum measurement results between different optimizers, they achieved a 100-fold improvement in efficiency compared to naive implementations. This technique is particularly valuable for computing properties across different physical parameters, such as energy surfaces for molecules at varying nuclear separations.

Despite its benefits, parameter sharing introduces challenges in the optimization landscape. Wang et al. \cite{Wang2023TrainabilityInitialization} showed that appropriate parameter initialization strategies are crucial when using shared parameters. By reducing the initial domain of each parameter inversely proportional to the square root of the circuit depth, they proved that the magnitude of the cost gradient decays at most polynomially with respect to the qubit count and circuit depth, enhancing trainability.

The implementation of parameter sharing in VQCs takes various forms, including explicit parameter sharing (directly constraining parameters to have identical values) and Bayesian optimization with information sharing. Each approach offers different trade-offs between optimization efficiency and circuit expressivity.

\subsection{The Barren Plateau Problem}\label{sec:related:barren}

The barren plateau phenomenon represents one of the most significant challenges in training variational quantum circuits. First identified by McClean et al. \cite{McClean2018BarrenLandscapes}, barren plateaus refer to regions in the optimization landscape where gradients vanish exponentially with the number of qubits, rendering gradient-based optimization ineffective for large-scale quantum systems.

Mathematically, for random parameterized quantum circuits (RPQCs), the variance of the gradient decreases exponentially with system size: $\text{Var}[\partial_k E] \propto 2^{-n}$, where $n$ is the number of qubits \cite{McClean2018BarrenLandscapes}. This exponential scaling means that the probability of measuring a non-zero gradient (to some fixed precision) becomes vanishingly small as the system size increases, requiring an exponential number of measurements to resolve gradients with sufficient accuracy.

The origin of barren plateaus can be understood through the lens of concentration of measure in high-dimensional spaces \cite{Cerezo2021CostCircuits}. As the dimension of Hilbert space grows exponentially with the number of qubits, the value of any reasonably smooth function tends to concentrate around its average with exponential probability. This concentration effect is particularly pronounced in circuits that form approximate 2-designs, which randomize quantum states in a way that mimics Haar-random unitaries \cite{Holmes2022ConnectingPlateaus}.

Several factors influence the emergence of barren plateaus, including circuit depth, architecture, and the locality of cost functions. Cerezo et al. \cite{Cerezo2021CostCircuits} demonstrated that using local cost functions (depending only on a subset of qubits) can mitigate barren plateaus, with gradients that decay at most polynomially rather than exponentially with system size. Similarly, Volkoff and Coles \cite{Volkoff2021LargeCircuits} showed that certain ansatz structures, particularly those with shallow depths or limited entanglement, can avoid barren plateaus.

Detection methods for barren plateaus include variance analysis (computing gradient variances across multiple random initializations), quantum optimal control tools (analyzing the controllability of the quantum system), and theoretical bounds on gradient variances based on circuit structure \cite{Arrasmith2021EffectOptimization}. These approaches allow practitioners to identify potential barren plateau issues before investing computational resources in training.

Mitigation strategies have evolved along several lines: structured ansatz design (using problem-inspired circuits instead of random ones), initialization strategies (such as identity block initialization or layer-wise pre-training), and adaptive optimization methods that do not rely solely on gradients \cite{Grant2019ANCircuits}. Weight re-mapping techniques have also shown promise in improving convergence for quantum variational classifiers \cite{Kolle2022ImprovingRe-Mapping}. The quantum natural gradient, which accounts for the geometry of the quantum parameter space, has also shown promise in navigating barren plateaus more effectively \cite{Stokes2020QuantumGradient}.

Despite these advances, the barren plateau problem remains a fundamental challenge in scaling quantum machine learning to larger systems, highlighting the need for continued research into circuit design and optimization strategies tailored to the unique characteristics of quantum systems.

\subsection{Deceptive Gradients in Quantum Machine Learning}\label{sec:related:deceptive}

While barren plateaus represent regions where gradients vanish exponentially, deceptive gradients present a complementary challenge: gradient information exists but does not reliably lead to the global optimum. This phenomenon is particularly relevant to the user's research on parameter sharing in variational quantum circuits, where the optimization landscape can become highly complex and misleading.

Deceptive gradients arise from several interrelated factors in quantum machine learning. First, the parameter sharing structure itself can create complex dependencies in the gradient information \cite{Wang2023TrainabilityInitialization}. When parameters are shared across multiple gates, the gradient with respect to a shared parameter becomes a sum of gradients from all gates using that parameter. These individual components may interfere destructively, leading to directions that appear promising but result in suboptimal solutions \cite{Self2021VariationalSharing}.

The Fourier structure of quantum circuits further contributes to gradient deception. Variational quantum circuits typically evaluate to finite Fourier series of the input parameters, creating a periodic structure with multiple equivalent minima and complex gradient patterns \cite{Wierichs2022GeneralGradients}. Parameter sharing can distort this Fourier structure, creating deceptive paths in the optimization landscape that lead to local minima rather than global optima.

Entanglement effects also play a crucial role in the emergence of deceptive gradients. High entanglement between qubits creates complex dependencies in the gradient information, and parameter sharing across entangling gates can lead to particularly misleading gradient directions \cite{OrtizMarrero2021Entanglement-InducedPlateaus}. The interplay between entanglement and parameter sharing creates optimization pathways that appear promising but lead to poor solutions, especially in deep circuits.

In practical implementations, measurement noise and finite sampling introduce additional layers of deception. Gradients are typically estimated through finite sampling, introducing noise that can make certain directions appear more promising than they actually are \cite{Arrasmith2021EffectOptimization}. Parameter sharing amplifies this effect by creating dependencies between noisy gradient estimates, potentially leading optimization algorithms astray. As we focus on experiments on classical simulators instead of real quantum hardware, we exclude the effects of noise in our analysis.

Several approaches have been proposed to address the challenge of deceptive gradients. Adaptive parameter sharing techniques dynamically adjust sharing patterns during optimization, starting with minimal sharing and gradually increasing it based on correlation analysis \cite{Skolik2021LayerwiseNetworks}. The quantum natural gradient, which accounts for the quantum geometry of the parameter space, provides more reliable gradient information that is less susceptible to deception \cite{Stokes2020QuantumGradient}. Recent work has also explored various optimization techniques specifically tailored for VQCs in reinforcement learning contexts \cite{Kolle2024ALearning}.

Hybrid optimization approaches combine gradient-based and gradient-free methods, using gradient-free techniques to escape deceptive regions and employing gradient information only in well-behaved areas of the landscape \cite{Arrasmith2021EffectOptimization}. Layerwise training strategies train the circuit layer by layer rather than all at once, reducing the complexity of the optimization landscape at each stage and minimizing the impact of deceptive gradients \cite{Skolik2021LayerwiseNetworks}.

Understanding and mitigating deceptive gradients remains an active area of research in quantum machine learning. As quantum hardware scales and more complex quantum models are developed, addressing the challenges posed by deceptive gradients will be crucial for realizing the potential advantages of quantum machine learning in practical applications.

\section{METHODOLOGY}\label{sec:method}

\subsection{Concept of Resolution}\label{sec:resolution}

In our experiments, we repeatedly sample the parameter space of the quantum circuits. We use the concept of resolution to describe the number of samples taken in the parameter space, interpreting the resolution as the granularity of our sampling.
As with most classical parameter optimization problems in machine learning, the parameters of quantum variational circuits are usually continuous. However, since the trainable parameters often describe rotations, they are usually periodic.
If we sample the parameter space of a quantum circuit with a resolution of 180, we divide the parameter space from $0$ to $4\pi$ into 180 equally spaced intervals, resulting in measurements every $4\pi/180 \approx 0.06981$.
As the resolution increases, the required number of measurements grows exponentially. For instance, a sub-circuit with three trainable parameters requires $90^3 = 7.29\times10^5$ measurements at a resolution of 90, while a higher resolution like $1440^3 \approx 3.99\times10^9$ leads to a significantly higher number of experiments.

\begin{figure}[htbp]
    \centerline{\includegraphics[width=\columnwidth]{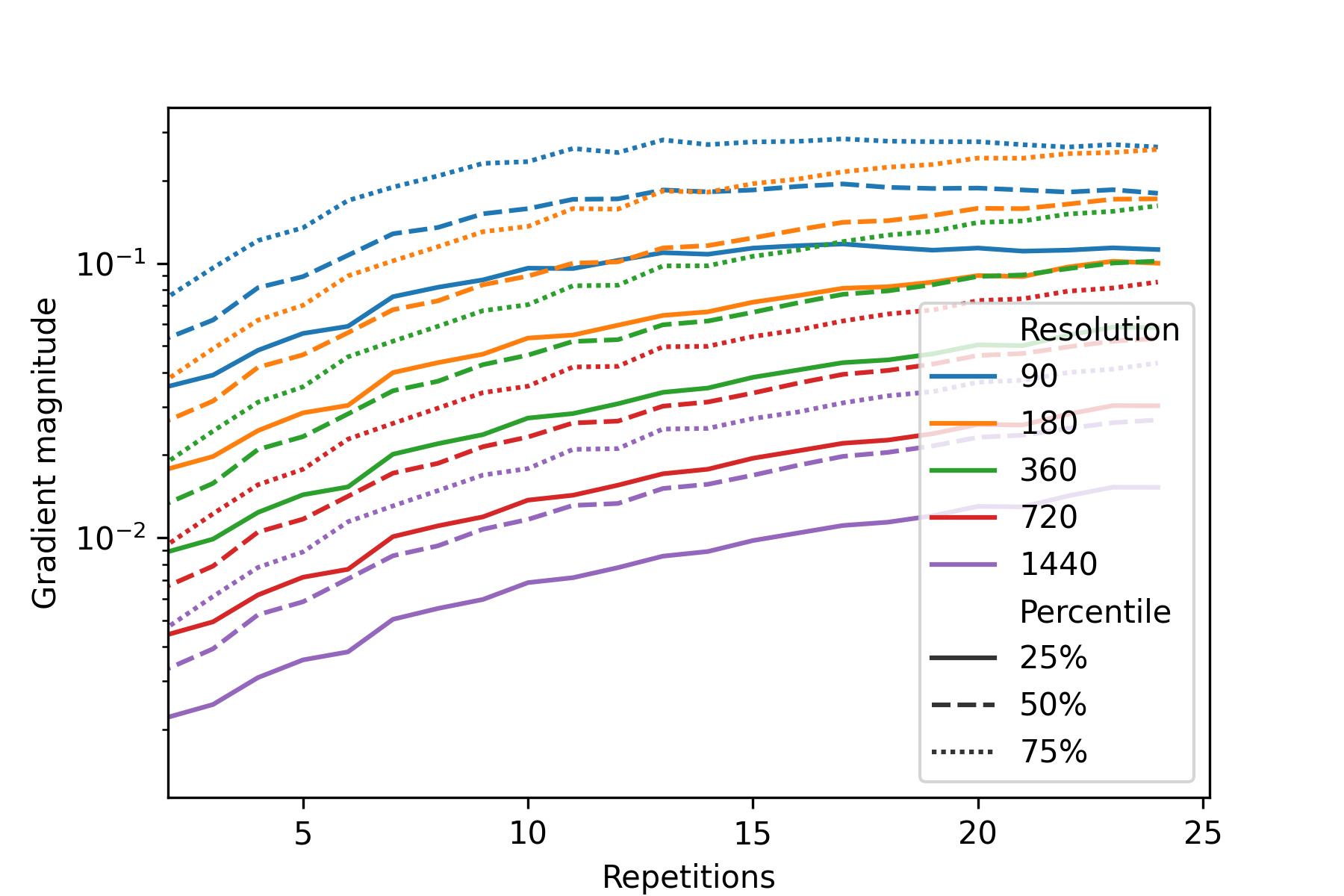}}
    \caption{Lower resolutions create scaling gradient magnitudes, shifting the distribution of gradients consistently.}
    \label{fig:resmagn}
\end{figure}

To mitigate this, we measure the impact of the resolution on the magnitude of the measured gradients (see \cref{fig:resmagn}).
From these experiments, we conclude that although the resolution influences the distribution of gradients, with higher resolutions leading to lower magnitudes on average, they scale similarly. Focusing on higher iterations, each halving of the resolution (and therefore doubling the coarseness) leads to an upward shift in the distribution of gradient magnitudes by about 25\% quantile (a more classical box-plot version is depicted in \cref{fig:resmagnv2}). This is visible in \cref{fig:resmagn}, where the 25\% quantile of the doubled resolution lays at the same position as the 50\% quantile of the lower resolution. The only notable exception is the very coarse resolution of only 90 sampling points (only included for reference), which does not maintain this scaling behavior.
We conclude that despite the resolution heavily influencing the distribution of gradients, due to the scaling, we can approximate the gradients of a circuit with a lower resolution. Doubling the resolution shifts the 2nd quantile of the distribution to the position of the 3rd quantile of the lower resolution. This scaling effect also translates to other quantiles and outliers (see \cref{fig:resmagnv2}).

\subsection{Deceptiveness}\label{sec:deceptiveness}
We use the concept of deceptiveness to measure the difficulty of an optimization landscape, particularly when impacted by parameter sharing in quantum circuits.
For each of our sampling points, we calculate the gradient of the cost function with respect to all parameters and then determine the direction of the gradient.

\begin{algorithm}
    \SetKwData{Validl}{valid\_l}
    \SetKwData{Validr}{valid\_r}
    \SetKwData{Validu}{valid\_u}
    \SetKwData{Validd}{valid\_d}
    \SetKwData{MaskOne}{M1}
    \SetKwData{Mask}{M}
    \SetKwData{GM}{opt}
    \SetKwData{Grad}{G}
    \SetKwData{Tol}{tol}
    \SetKwData{Val}{V}
    \SetKwData{gradtol}{tol\_g}
    \SetKwFunction{Where}{where}
    \SetKwFunction{Min}{min}
    \SetKwFunction{Roll}{roll}
    \SetAlgoLined
    \KwIn{\textit{values} \Val $\in \mathbb{R}^{N \times N}$,
        \textit{gradients} \Grad $\in \mathbb{R}^{N \times N \times 2}$,
        \textit{tolerance} \Tol,
        \textit{gradient tolerance} \gradtol
    }
    \KwOut{Mask \Mask $\in \{-1,0,1\}^{N \times N}$}

    \Mask $\gets -1$ (matrix of same shape as $V$)\;
    \GM $\gets$ (\Val $ - $ \Min(\Val) ) $<$ \Tol\;
    \Mask $\gets $ \Where(\GM, 1, \Mask)\;

    \Validl $\gets$ \Grad$[:,:,0]$ $\geq -$\gradtol\;
    \Validr $\gets$ \Grad$[:,:,0]$ $\leq +$\gradtol\;
    \Validu $\gets$ \Grad$[:,:,1]$ $\geq -$\gradtol\;
    \Validd $\gets$ \Grad$[:,:,1]$ $\leq +$\gradtol\;

    \Repeat{\Mask does not change}{
        \MaskOne $\gets$ \Mask == 1 (Mask of currently non-deceptive points)\;
        \Mask $\gets$ \Where(\Validl $\land$ \Roll(\MaskOne, +1, 0), 1, \Mask)\;
        \Mask $\gets$ \Where(\Validr $\land$ \Roll(\MaskOne, -1, 0), 1, \Mask)\;
        \Mask $\gets$ \Where(\Validu $\land$ \Roll(\MaskOne, +1, 1), 1, \Mask)\;
        \Mask $\gets$ \Where(\Validd $\land$ \Roll(\MaskOne, -1, 1), 1, \Mask)\;
    }
    \Mask $\gets$ \Where(\GM, 0, \Mask)\;
    \Return \Mask\;
    \caption{Gradient Deceptiveness Detection Algorithm. This algorithm identifies points in the parameter space that lead to global minima versus those that are deceptive (leading to local minima). Initially, we mark all global minima within tolerance. Then, we iteratively propagate this information along gradient directions, marking points that would flow toward global minima when following gradient descent. This approach is particularly important when analyzing quantum circuits with parameter sharing, as shared parameters create complex dependencies in the optimization landscape. The gradient tolerance allows for handling numerical imprecision by treating very small gradients (e.g., $\leq 10^{-7}$) as effectively zero.}\label{alg}
\end{algorithm}

We define the deceptiveness of a sampling point using a binary mask, as described in Algorithm \ref{alg}. Each entry in the mask indicates whether following the gradient with a very small step size would lead to a global minimum or just a local minimum. Since we are using different resolutions, we choose to accept local minima as global minima if they are within a certain tolerance \verb|tol| of the global minimum. This approach ensures that we do not miss a global optimum due to not sampling its direct vicinity.

Parameter sharing in quantum circuits introduces additional complexity to the optimization landscape. When parameters are shared across different gates, changing one parameter affects multiple parts of the circuit simultaneously, creating interdependencies that can significantly alter gradient behavior. Our algorithm detects these effects by propagating information about optimization paths through the parameter space, identifying regions where parameter sharing creates deceptive gradients that trap optimizers in suboptimal solutions.

\begin{figure}[htbp]
    \centerline{\includegraphics[width=\columnwidth]{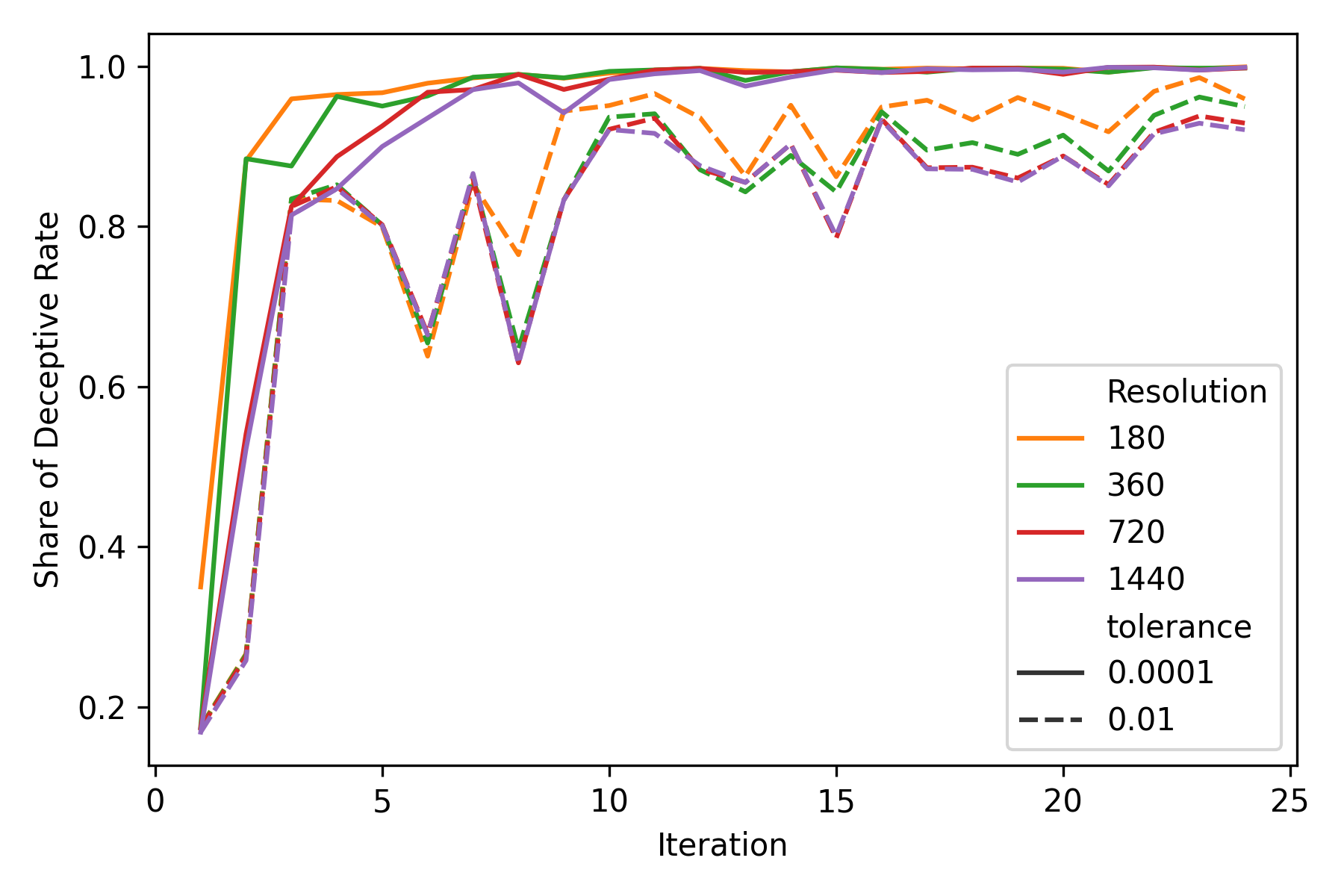}}
    \caption{Lower resolutions create roughly the same deceptiveness.}
    \label{fig:resdecept}
\end{figure}

We then calculate the deceptiveness of the solution sampling by determining the ratio of deceptive points to the total number of points (with global minima not counted as deceptive). As shown in \cref{fig:resdecept}, different resolutions lead to roughly the same deceptiveness ratios. For larger tolerances like $10^{-2}$ (marked with solid lines in \cref{fig:resdecept}), the deceptiveness at different coarseness levels is nearly identical. For smaller tolerances like $10^{-4}$ (shown with a dashed line in \cref{fig:resdecept}), where the number of local minima accepted as global minima is much smaller, the deceptiveness ratio is still similar, demonstrating consistent scaling effects, with only a few outliers, where the highest resolution found optima the others have skipped.
Therefore, we conclude that the deceptiveness of a sampling point is not significantly influenced by the resolution. This allows us to use lower resolutions for all experiments hereinafter (360 if not defined otherwise) while still accurately capturing the effects of parameter sharing on optimization trajectories.
We have shown that this holds for higher number of qubits in \cref{sec:resolutionhigher} as well.

\subsection{Experiments}\label{sec:experiments}

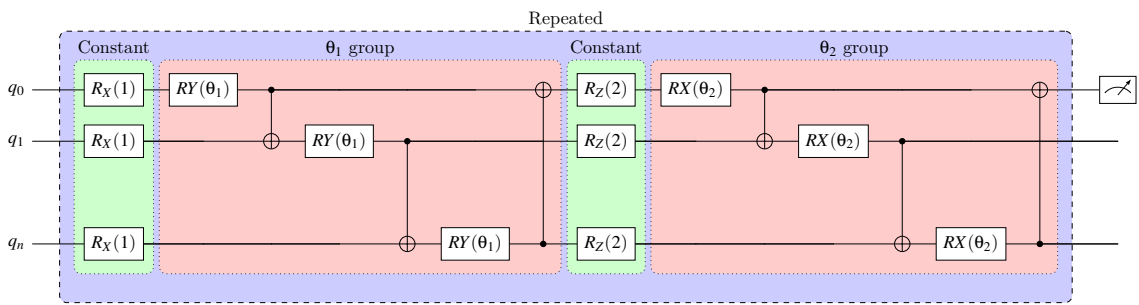
\begin{figure*}[!h]
    \centering
    \begin{adjustbox}{width=2\columnwidth}
        \begin{quantikz}[row sep={1.0cm,between origins },wire types={q,q,q},thin lines]
            \lstick{$q_0$} &[0.5cm] \gate{R_X(1)} \gategroup[3,steps=14,style={dashed,rounded corners,fill=blue!20, inner xsep=10pt, inner ysep=20pt},background]{{Repeated}} \gategroup[3,steps=1,style={dotted,rounded corners,fill=green!20, inner xsep=2pt},background]{{Constant}} & \gate{RY(\theta_1)} \gategroup[3,steps=6,style={dotted,rounded corners,fill=red!20, inner xsep=2pt},background]{{$\theta_1$ group}} & \ctrl{1} & \qw                 & \qw      & \qw                 & \targ{}   & \gate{R_Z(2)} \gategroup[3,steps=1,style={dotted,rounded corners,fill=green!20, inner xsep=2pt},background]{{Constant}} & \gate{RX(\theta_2)} \gategroup[3,steps=6,style={dotted,rounded corners,fill=red!20, inner xsep=2pt},background]{{$\theta_2$ group}} & \ctrl{1} & \qw                 & \qw      & \qw                 & \targ{}   &[0.5cm]  \meter{} \\
            \lstick{$q_1$} &      \gate{R_X(1)}                                                                                                                                                                                                                                       & \qw                                                                                                                                 & \targ{}  & \gate{RY(\theta_1)} & \ctrl{1} & \qw                 & \qw       & \gate{R_Z(2)}                                                                                                           & \qw                                                                                                                                 & \targ{}  & \gate{RX(\theta_2)} & \ctrl{1} & \qw                 & \qw       &  \qw      \\[1cm]
            \lstick{$q_n$} &      \gate{R_X(1)}                                                                                                                                                                                                                                       & \qw                                                                                                                                 & \qw      & \qw                 & \targ{}  & \gate{RY(\theta_1)} & \ctrl{-2} & \gate{R_Z(2)}                                                                                                           & \qw                                                                                                                                 & \qw      & \qw                 & \targ{}  & \gate{RX(\theta_2)} & \ctrl{-2} &  \qw
        \end{quantikz}
    \end{adjustbox}
    \caption{The default circuit used in our experiments. }
    \label{fig:circ:default}
\end{figure*}

In our experimental framework, unless explicitly stated otherwise, we employed the default quantum circuit depicted in \cref{fig:circ:default}. A comprehensive explanation of this circuit and its configuration is provided in \cref{sec:defaultcircuit}.

In our experiments,  we will run the Adam optimizer \cite{Kingma2014Adam:Optimization} and the Stochastic Gradient Descent (SGD) optimizer \cite{Robbins1951AMethod} on variational quantum circuits with different number of parameter sharing applications, reusing the same parameter combination up to 20 times, to evaluate their performance in the context of deceptiveness.
To allow both optimizers to live up to their full potential, we execute both optimizers with different learning rates (0.0001, 0.001, 0.01, 0.1, 1.0) on 200 uniformly sampled initial points in the parameter space. We allow for 500 iterations of each configuration, tracking intermediate steps and losses. 
As before, we utilize our default circuit (\cref{sec:defaultcircuit}) with two parameters, with the loss function motivating the optimization process to find the smallest possible value of the output of the circuit. 
We use a high resolution (1440 sampling points per dimension) to identify the true global minima, defining this as the ground truth per number of parameter sharing applications.

We then compare the performance of both optimizers in relation to the measured deceptiveness ratios determined in \cref{sec:deceptiveness}, showing that higher number of parameter sharing applications lead to higher deceptiveness ratios, which in turn leads to significantly worse performance of the optimizers by showing both selected trajectories and minimal achieved losses per configuration.

\section{RESULTS}\label{sec:results}

\subsection{Plot overview}\label{sec:plotoverview}

\begin{figure}[htbp]
    \centerline{\includegraphics[width=\columnwidth]{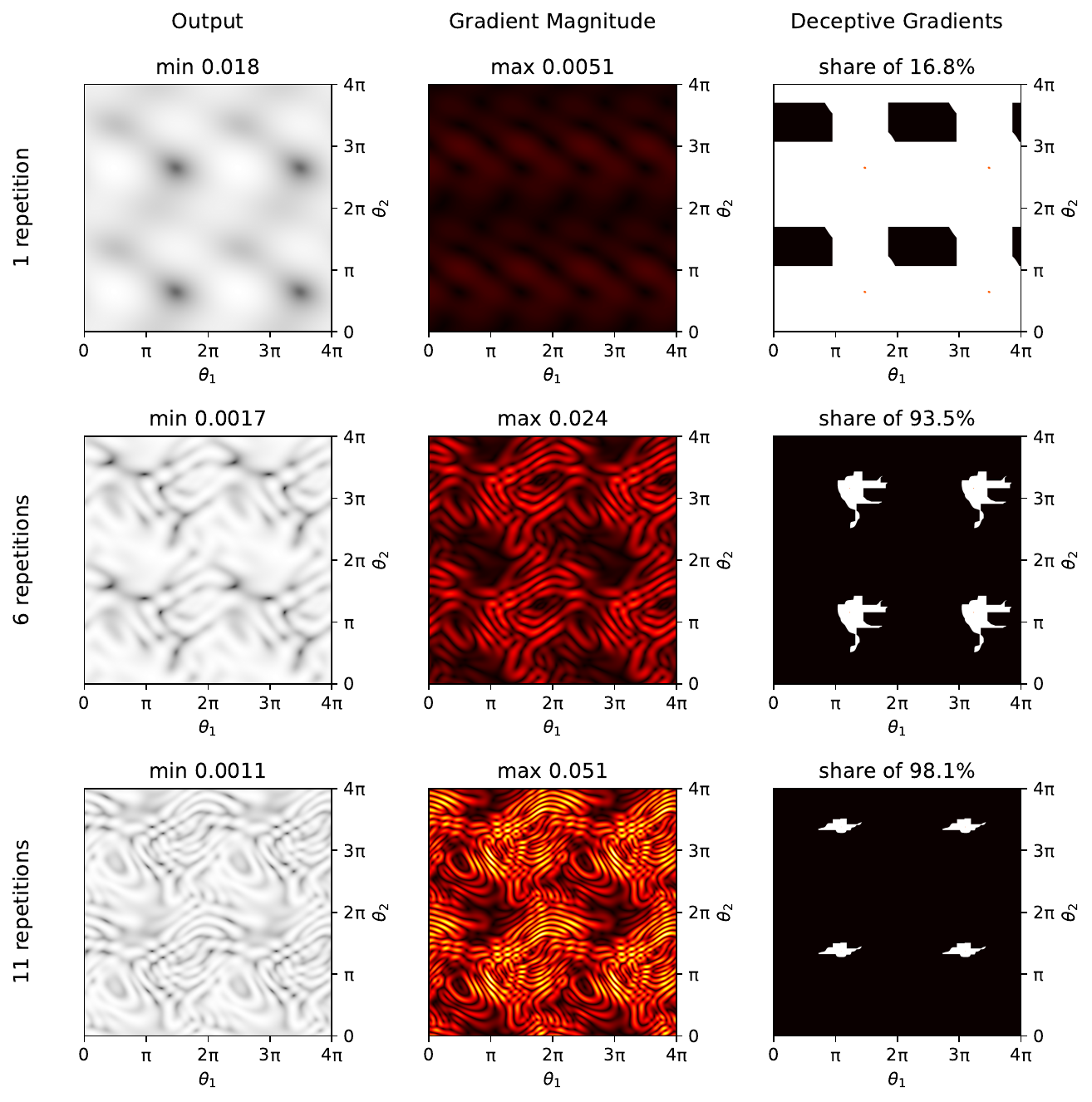}}
    \caption{Exemplary run of circuit deceptiveness at different parameter sharing applications.}
    \label{fig:3x3}
\end{figure}

We will first introduce the plots we will be using in the following sections. In \cref{fig:3x3} we showcase an exemplary run on our default circuit. Each row shows a different number of parameter sharing applications, with the columns showing circuit output, gradient magnitude and deceptiveness masks. The output image displays the value of the circuit output (which we treat as a loss function, e.g., the target to minimize) for each parameter combination. Darker areas indicate more favorable parameter combinations (the colors are distributed exponentially, not linearly). The smallest measured value is listed over each output image. The second column shows the gradient magnitude of the circuit output with respect to the parameters, which more intense red indicating a higher gradient magnitude at that parameter combination, with the peak of each gradient magnitude being listed over the image. The third column shows the deceptiveness mask, e.g., the mask we calculate in \cref{sec:deceptiveness} for each parameter combination, with bright areas indicating areas pointing towards a favorable direction and dark areas showing deceptive areas, with the share of deceptive areas listed over the image. 

From the \cref{fig:3x3}, we can derive first observations (note that these are only showcases and a numeric analysis will be done in \cref{sec:resolution}): higher number of parameter sharing applications lead to a better global minima (and are therefore more favorable), but this also leads to a significantly more complex solution landscape (complex shapes in the first column) and higher gradient magnitudes (more red in the second column), which in turn leads to a higher deceptiveness ratio (more dark areas in the third column).

\subsection{Optimizer trajectories}\label{sec:trajectories}
\begin{figure}[htbp]
    \centerline{\includegraphics[width=\columnwidth]{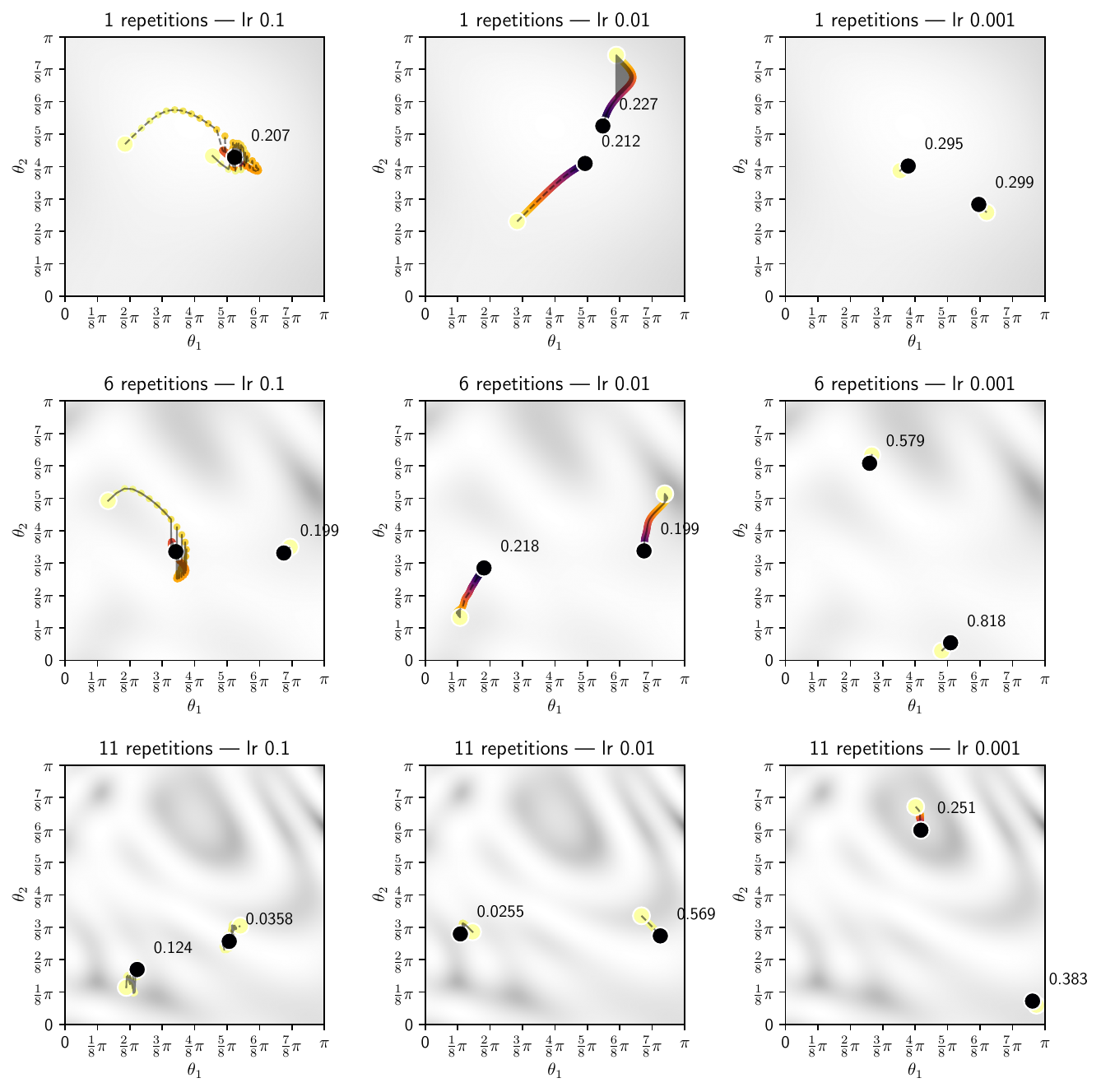}}
    \caption{Different trajectories of the Adam optimizer on the default circuit with different parameter sharing applications and learning rates.}
    \label{fig:trajectories:adam}
\end{figure}

In the following, we analyze optimizer trajectories with a primary focus on the Adam optimizer.
In \cref{fig:trajectories:adam}, we present selected trajectories of the Adam optimizer on circuits with one, six, and eleven parameter sharing applications (depicted in the different rows) and learning rates of $0.1$, $0.01$, and $0.001$ in the columns. We focus on parameters in the range of $[0, \pi]$ (a numerical analysis on the full parameter space is shown in \cref{sec:successrate}), essentially the lower left corner of \cref{fig:3x3}. To minimize visual clutter, we only plot the initial steps until convergence or stagnation. The trajectory is colored brighter at the beginning and darker at the end, with both start and end points highlighted. Some notable endpoints are annotated with their corresponding loss values.

Evaluating the trajectories in \cref{fig:trajectories:adam}, we observe that for the circuit with one parameter sharing application (i.e., no actual parameter sharing) in the first row, the optimizer with higher learning rates converges rather quickly near the global optimum (the dark area roughly in the middle of the image). For a lower learning rate (first row, second column), the optimizer begins to exhibit signs of struggle: the black area in the top right corner results from the optimizer repeatedly traversing a gradient ridge, causing the trajectory to oscillate. The lowest learning rate (first row, third column) demonstrates minimal movement.

Moving to the second row, the learning rate of $0.1$ exhibits a combination of the previously observed behaviors: for one starting point, it converges to a local minimum, while for another starting point in a barren plateaus region with no meaningful gradient, it barely moves at all. Notably, both seeds fail to find the global minimum (consistently marked in dark gray). The learning rate of $0.01$ (second row, second column) achieves better performance, successfully traversing the barren plateaus region and converging to a local minimum.

For the highest number of repeated parameter sharing applications (eleven, third row), none of the configurations is able to converge to a meaningful optimal solution.

We therefore conclude that different numbers of parameter sharing applications necessitate different optimal learning rates for optimizers, with higher numbers of parameter sharing applications creating more complex solution landscapes and leading to lower success rates across all optimizers.

\subsection{Optimizer Success Rate}\label{sec:successrate}

\begin{figure}[htbp]
    \centerline{\includegraphics[width=\columnwidth]{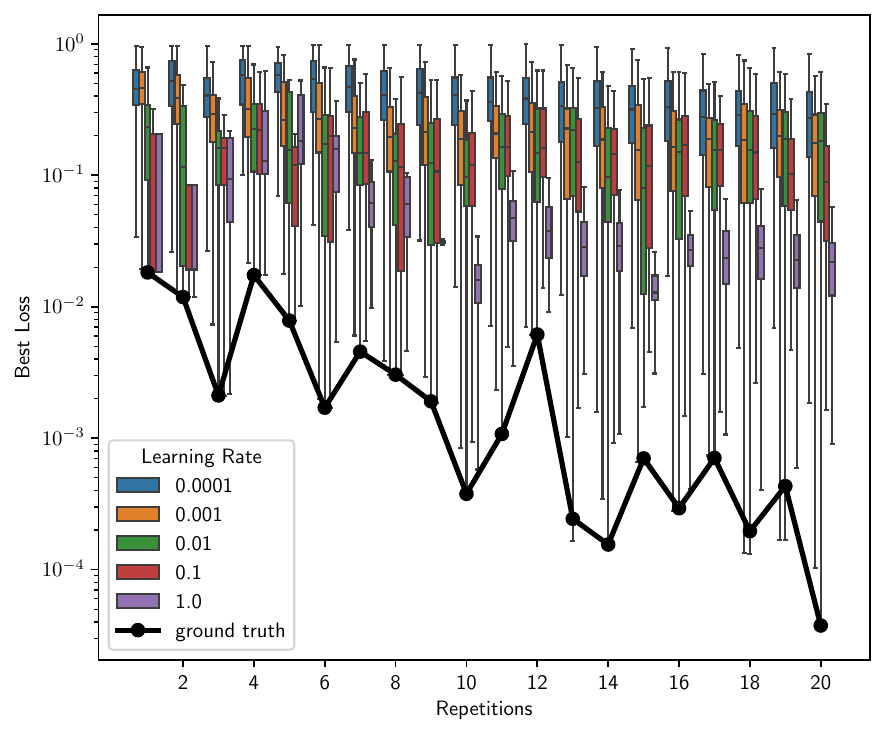}}
    \caption{Distribution of the minimal loss of the SGD optimizer measured during each run for different parameter sharing applications and learning rates, compared to the global minimum and deceptiveness ratio.}
    \label{fig:optsgd_success_box}
\end{figure}

We analyze the achieved loss values of the Adam and SGD optimizers against the global minimum and deceptiveness ratio. For the SGD optimizer, we monitor the distribution of minimal losses across multiple runs, as depicted in \cref{fig:optsgd_success_box}. Here, we focus solely on the best loss achieved, ignoring potential later degradation in performance, effectively simulating elitism. The ground truth (thick black line) represents the global minimum, estimated at high resolution. From the bar plot, we observe that this estimated curve is rarely undercut by the optimizers and generally aligns precisely with the lowest loss achieved, thereby serving as a reliable reference point. Across all configurations, lower learning rates perform worse for stochastic gradient descent. It is noteworthy, however, that the lowest achieved loss values are occasionally produced by these low rates (0.001 to 0.01), though these represent outliers in the distribution.

\begin{figure}[htbp]
    \centerline{\includegraphics[width=\columnwidth]{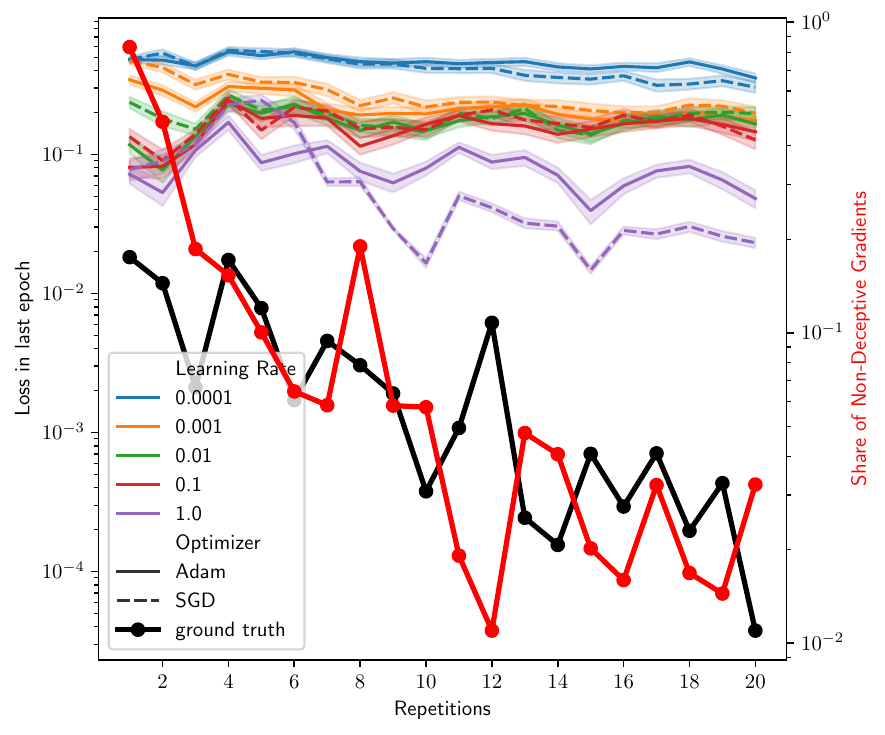}}
    \caption{Minimal loss of the Adam and SGD optimizers measured during each run for different parameter sharing applications and learning rates, compared to the global minimum and deceptiveness ratio.}
    \label{fig:optadamsgd_sucess_bestloss}
\end{figure}

When examining the mean loss of the SGD and Adam optimizers in \cref{fig:optadamsgd_sucess_bestloss}, we observe that learning rates performing poorly on average (such as the very low 0.0001 for both optimizers) are relatively unaffected by the number of parameter sharing applications. Better-performing learning rates achieve improved results with higher numbers of parameter sharing applications, with the SGD optimizer demonstrating notably superior performance compared to the Adam optimizer. It is important to note, however, that neither optimizer consistently matches the actual global minimum (only some outliers achieve this, as depicted in \cref{fig:optsgd_success_box}).
We also observe that the proportion of non-deceptive gradients decreases with higher numbers of repetitions, which aligns with our findings in \cref{sec:deceptiveness}.

This increased deceptiveness prevents the optimizers from fully capitalizing on the improved loss potential, yielding only up to one order of magnitude in performance improvement instead of the potential four orders of magnitude. It is also significant to note that the surprisingly good performance of the SGD optimizer at the highest learning rate is not directly attributable to the quality of the solution algorithm but rather to its high nondeterminism. Since we analyze only the best loss achieved at any point during a run, the optimizer demonstrates highly explorative behavior, effectively escaping local minima through substantial parameter jumps. This approach, however, leads to second-order effects where the optimizer can escape local minima but also overshoot the global minimum, ultimately degrading performance (see \cref{sec:additionaloptimizer} and \cref{fig:optsgdadam_sucess_lastepoch}).

\section{CONCLUSION}\label{sec:conclusion}

Our research has yielded several significant insights into the optimization landscape of variational quantum circuits. We have demonstrated that parameter sharing creates a complex trade-off in these circuits: while it enhances circuit expressivity and can lead to superior global minima by up to multiple orders of magnitude, it simultaneously increases landscape deceptiveness. This fundamental tension underlies many of the challenges in quantum circuit optimization.

Our methodological contributions include establishing that the deceptiveness ratio of quantum circuit optimization landscapes remains largely independent of the resolution parameter. This finding enables more efficient analysis using lower resolutions (360 sampling points) without compromising the integrity of results. To support this analysis, we developed and validated a novel gradient deceptiveness detection algorithm that effectively identifies points in the parameter space that lead to either global or local minima when following gradient descent paths.

The experimental results reveal a clear pattern: higher numbers of parameter sharing applications generate significantly more complex solution landscapes with increased gradient magnitudes, which translates to higher deceptiveness ratios. This increased deceptiveness has a measurable impact on optimizer performance, particularly for gradient-based methods such as Adam and SGD, which show degraded convergence as parameter sharing increases. Furthermore, our analysis confirms that while the barren plateau problem scales with qubit count, the relative deceptiveness patterns remain consistent across different resolution configurations for quantum systems up to four qubits.

An important practical observation from our work is that optimizer learning rate selection becomes increasingly critical in highly deceptive landscapes. High learning rates occasionally enable escape from local minima, though this comes at the cost of potential solution stability. The methodology we have introduced provides a framework for quantitatively measuring optimization difficulty in quantum circuits, which is particularly valuable when evaluating the impact of circuit design choices on trainability.

Perhaps most significantly, our findings suggest a fundamental mismatch between traditional optimization approaches and quantum machine learning landscapes. While gradient-based optimizers like Adam excel in classical machine learning contexts, they struggle substantially with quantum machine learning landscapes. This points to the need for gradient-free optimization approaches such as evolutionary algorithms for more effective quantum circuit training.

Building on these findings, several promising directions for future research emerge. First, we aim to extend our analysis to evaluate the performance of evolutionary algorithms as potential alternatives to gradient-based optimizers for highly deceptive quantum landscapes. This would include comprehensive benchmarking of genetic algorithms, particle swarm optimization, differential evolution, and covariance matrix adaptation evolution strategies (CMA-ES).

Additionally, there is significant potential in investigating adaptive optimization strategies that can dynamically adjust parameter sharing based on landscape deceptiveness detected during training. This approach could help balance the trade-off between expressivity and trainability. We also plan to explore the impact of circuit depth and entanglement patterns on deceptiveness metrics to develop circuit design guidelines that similarly balance these competing considerations.

To address practical implementation concerns, we intend to extend the analysis to noisy quantum simulations to evaluate how hardware errors interact with parameter sharing and gradient deceptiveness. Finally, we see considerable promise in developing hybrid optimization approaches that combine the strengths of gradient-based and gradient-free methods to navigate complex quantum optimization landscapes more effectively. Such hybrid approaches may offer the best path forward for practical quantum machine learning applications.

\section*{\uppercase{Acknowledgements}}
This paper was partially funded by the German Federal Ministry of Education and Research through the funding program ``Quantum Computing User Network'' (contract number 13N16196).
This research is part of the Munich Quantum Valley,
which is supported by the Bavarian state government
with funds from the Hightech Agenda Bayern Plus.

\bibliographystyle{apalike}
{\small
\bibliography{bibliography}}

\section{APPENDIX}
\subsection{Default Circuit}\label{sec:defaultcircuit}
We employ the default parametrized circuit shown in \cref{fig:circ:default}. The circuit uses two variational parameters, $\theta_1$ and $\theta_2$, and consists of repeated blocks that combine fixed single-qubit rotations with two parameterized layers. Each block applies a constant $R_X(1)$ on all qubits, followed by a layer of $R_Y(\theta_1)$ rotations and a nearest-neighbor CNOT pattern with cyclic boundary conditions. It then applies a constant $R_Z(2)$ and a second parameterized layer of $R_X(\theta_2)$ rotations with the same entangling pattern. The circuit output is the probability of measuring $\ket{1}$ on the first qubit.

We select this architecture because it is a standard motif in circuit-centric quantum classifiers \cite{Schuld2020Circuit-centricClassifiers} and scales to higher qubit counts without changing connectivity. The fixed rotations act as controlled circuit structure and provide a consistent baseline when comparing parameter sharing settings.

\subsection{Resolution Analysis for Higher Qubit Counts}\label{sec:resolutionhigher}
In \cref{sec:resolution} and \cref{sec:deceptiveness}, we show that the deceptiveness ratio is largely insensitive to the sampling resolution for two-qubit circuits. Since barren plateaus intensify with system size \cite{McClean2018BarrenLandscapes}, we extend this analysis to higher qubit counts using the default circuit (\cref{fig:circ:default}, \cref{sec:defaultcircuit}).

\begin{figure}[h!]
    \centerline{\includegraphics[width=\columnwidth]{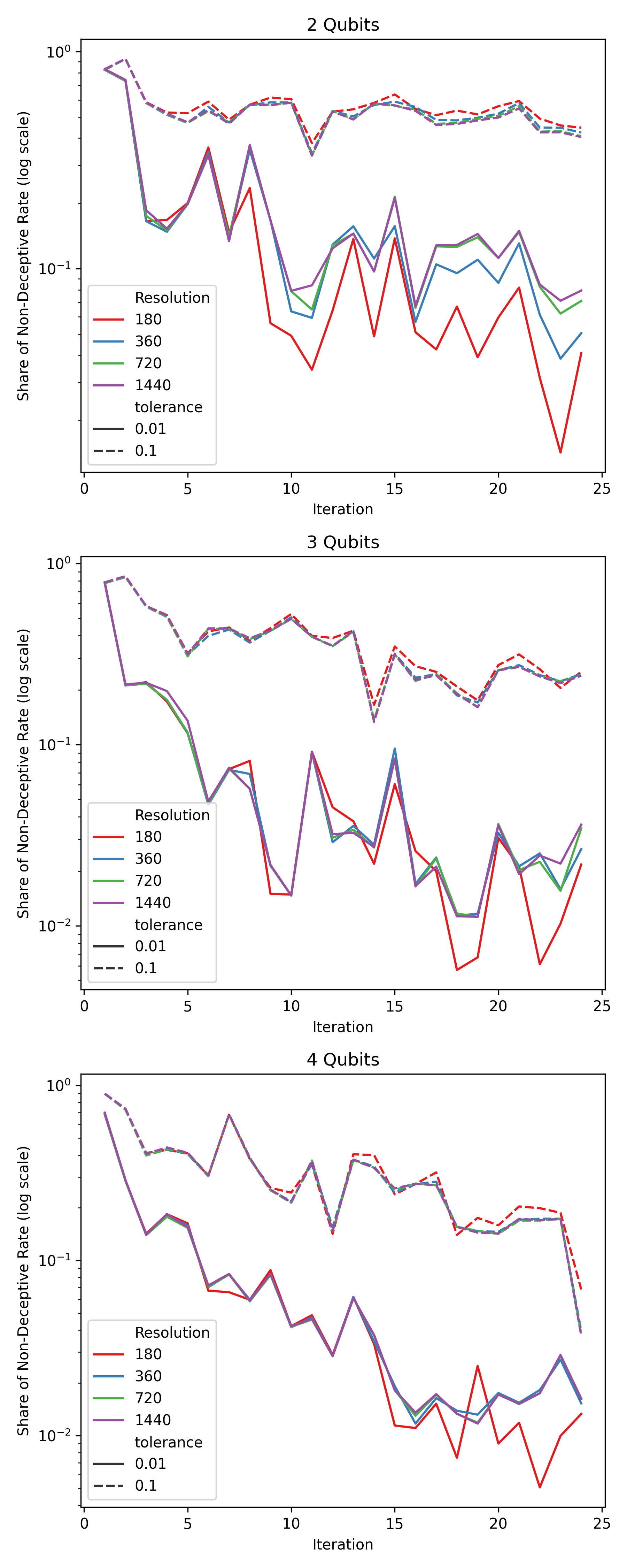}}
    \caption{Stability of deceptiveness metrics across varying resolutions persists in quantum systems with increasing qubit counts.}
    \label{fig:resdecept3}
\end{figure}

We evaluate resolutions of 180, 360, 720, and 1440 sample points per parameter and compute the proportion of non-deceptive gradients, where a gradient is classified as non-deceptive if infinitesimal descent converges to a global minimum or to a local minimum within a tolerance of $0.01$ or $0.1$ from the global optimum. \Cref{fig:resdecept3} summarizes the results for two, three, and four qubits.

Across all qubit counts, the deceptiveness ratio remains stable for resolutions of 360 and above. The main deviation occurs at the lowest resolution (180), where coarse sampling reduces the number of acceptable minima under strict tolerances.

\begin{figure}[h!]
    \centerline{\includegraphics[width=\columnwidth]{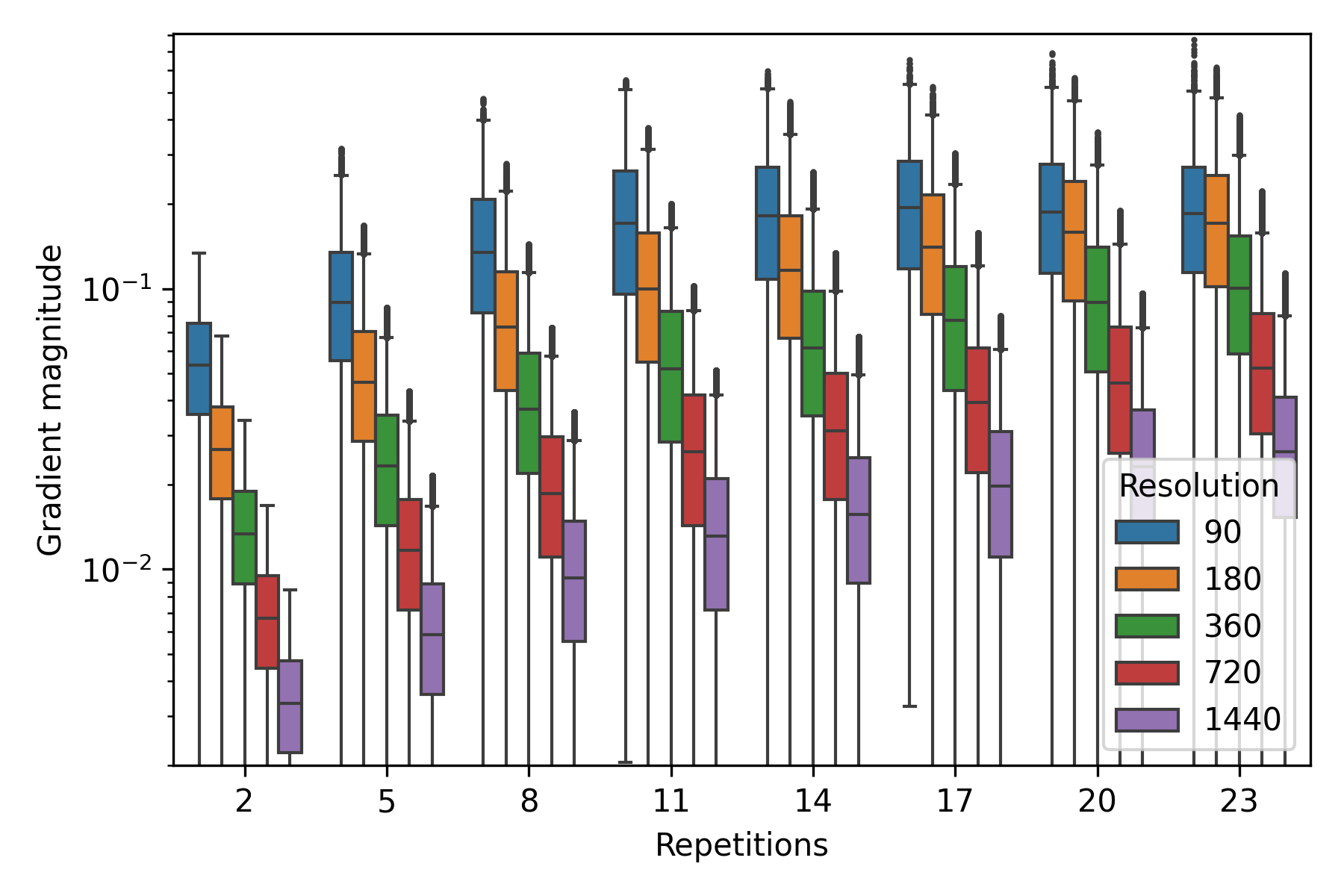}}
    \caption{Empirical distribution of gradient magnitudes across varying resolution configurations for a two-qubit quantum circuit. The observed scaling behavior aligns with theoretical predictions outlined in \cref{sec:resolution}, except the lowest resolution configuration (90 sample points), depicted in blue.}
    \label{fig:resmagnv2}
\end{figure}

\Cref{fig:resmagn} and \cref{fig:resmagnv2} complement this analysis by characterizing how resolution shifts the gradient-magnitude distribution while preserving the scaling behavior described in \cref{sec:resolution}.

Using resolution 1440 as a reference, resolutions of 360 and 720 yield negligible deviations in deceptiveness ratios across all tested qubit counts. These results support using moderate resolutions for landscape analysis in larger systems without materially affecting the deceptiveness estimates.

\subsection{Additional Optimizer Rates}\label{sec:additionaloptimizer}

\begin{figure}[h!]
    \centerline{\includegraphics[width=\columnwidth]{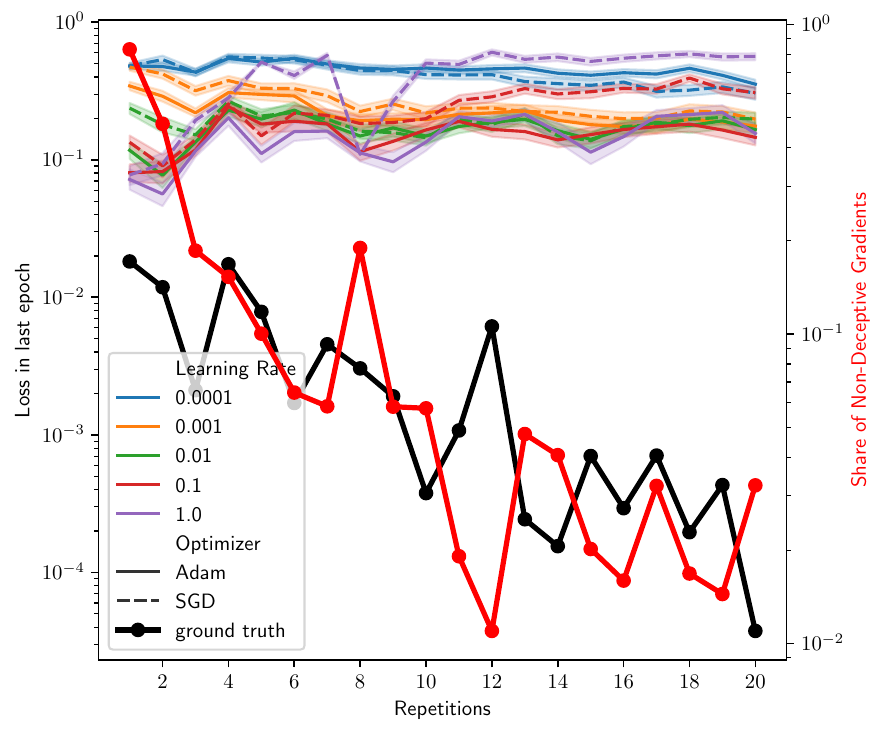}}
    \caption{Loss of the Adam and SGD optimizers at the last epoch for different parameter sharing applications and learning rates, compared to the global minimum.}
    \label{fig:optsgdadam_sucess_lastepoch}
\end{figure}

In \cref{fig:optsgdadam_sucess_lastepoch}, we report the final-epoch loss of Adam and Stochastic Gradient Descent (SGD) for different learning rates and parameter sharing settings. Compared to best-epoch reporting, the largest differences occur at high learning rates, where overshooting can reduce transient losses while degrading the final iterate. This effect is more pronounced for SGD, which becomes increasingly exploratory at very high learning rates in deceptive landscapes.

\end{document}